\pdfoutput=1

\documentclass[11pt]{article}

\usepackage[preprint]{acl}
\usepackage{ulem}
\usepackage{times}
\usepackage{latexsym}

\usepackage[T1]{fontenc}

\usepackage[utf8]{inputenc}

\usepackage{microtype}
\usepackage{url}
\usepackage{inconsolata}

\usepackage{graphicx}
\usepackage[most]{tcolorbox}
\usepackage{verbatim}
\usepackage[ruled,vlined]{algorithm2e}
\usepackage{amsmath}
\usepackage{booktabs}

\title{Dyve: Thinking Fast and Slow for Dynamic Process Verification}

\author{
  \textbf{Jianyuan Zhong\thanks{\ \ Equal contribution.}}, 
  \textbf{Zeju Li\footnotemark[1]}, 
  \textbf{Zhijian Xu}, 
  \textbf{Xiangyu Wen}, 
  \textbf{Qiang Xu\thanks{\ \ Corresponding author.}} 
  \\
  The Chinese University of Hong Kong \\
  \texttt{\{jyzhong, zjli24, zjxu21, xywen22, qxu\}@cse.cuhk.edu.hk}
}

\begin{document}
\maketitle
\begin{abstract}

We present \textbf{\textit{Dyve}}, a dynamic process verifier that enhances reasoning error detection in large language models by integrating fast and slow thinking, inspired by Kahneman's Systems Theory. Dyve adaptively applies immediate token-level confirmation (\textit{System 1}) for straightforward steps and comprehensive analysis (\textit{System 2}) for complex ones. Leveraging a novel step-wise consensus-filtered process supervision technique, combining Monte Carlo estimation, LLM-as-a-Judge, and specialized reasoning models, we curates high-quality supervision signals from noisy data for Dyve. Experimental results on ProcessBench and the MATH dataset confirm that Dyve significantly outperforms existing process-based verifiers and boosts performance in Best-of-N settings. Our code, data and model are released at: \url{https://github.com/staymylove/Dyve}

\end{abstract}

\section{Introduction}

Large Language Models (LLMs) have significantly enhanced their reasoning capabilities by shifting from rapid, intuitive System 1 responses to more deliberate, extended System 2 thinking~\cite{team2025kimik1.5, arrieta2025earlyexternalsafetytestingO3, guo2025deepseekr1}. While enabling more complex problem-solving in math and scientific reasoning, this has also introduced new challenges in process verification, particularly in the reliable evaluation of incomplete reasoning traces. 

Process-based verifiers (PRMs) are essential for detecting process errors. However, becuase human annotations for process supervision~\cite{Lightman2023LetsVS800k} are prohibitively expensive, researchers increasingly use Monte Carlo estimation methods~\cite{wang2024mathshepherd, luo2024improve_omegaprm} to annotate process labels, even though these labels are noisy and weak~\cite{zhang2025lessons}. Moreover, most verifiers rely on a simplistic "System 1" binary yes/no prediction, which is insufficient for capturing complex process errors.

Recently released reasoning LLMs, such as OpenAI O1~\cite{jaech2024openaio1} and DeepSeek R1~\cite{guo2025deepseekr1}, show promise in detecting process errors through reinforcement learning. Their reasoning traces include metacognitive cues (e.g. `hmm', `wait, let's check') that hint at a rudimentary verification mechanism, a kind of `aha' moment. However, since process verification was not the primary design goal, these abilities can be unreliable. Moreover, their reliance on a System 2–style self-correction process often leads to overthinking~\cite{chen2025overthinkingo1like} and reduce efficiency.

Our work introduces \textbf{\textit{Dyve}} (\textbf{Dy}namic Process \textbf{Ve}rifier), a specialized reasoning language model that dynamically detects process errors using fast and slow thinking, inspired by Kahneman's Systems Theory~\cite{kahneman_thinking_2012}. For reasoning traces from step 1 to $t$, Dyve adaptively applies either \textbf{\textit{System 1}}, which supplies single-token confirmation for clearly correct steps, or \textbf{\textit{System 2}} for deeper analysis to complex ones. 
To support this adaptive mechanism, we introduce a novel \textbf{\textit{step-wise consensus-filtered process supervision}} technique. Our method leverages Monte Carlo estimation to generate multiple rollouts per query, uses an LLM-as-a-Judge~\cite{Gu2024ASOJudge} to assess the full reasoning trace, and employs a reasoning LLM for step-by-step analysis to flag steps that require further verification. In doing so, we curate approximately 117K high-quality training examples from 1.2M noisy Monte Carlo rollouts, demonstrating that quality, not quantity, is key to effectively train an process-based verifier.

Experimental results on ProcessBench~\cite{zheng2024processbench} show that Dyve significantly outperforms existing PRMs and other reasoning models in detecting process errors in complete or incomplete reasoning traces. Furthermore, when combined with a proposer language model, Dyve yields better performances under Best-of-N then other PRMs.



\section{Related Work}

Recent research~\cite{setlur2024rewardingprogressscalingautomated,wang2024mathshepherd,guan2025rstar} shows that external reward models can improve LLM reasoning by selecting the best path from multiple candidates. Outcome Reward Models (ORMs)~\cite{cobbe2021ORM,yang2024qwen2.5math} optimize for final outputs but overlook vital intermediate steps. Process Reward Models (PRMs)~\cite{lightman2023lets_verify,zhang2025lessons,wang2024mathshepherd} provide rapid binary validations for each step, yet struggle with a deeper analysis of incomplete traces. In contrast, Generative Verifiers (GenRMs)~\cite{zhang2024generativeverifiersrewardmodeling} combine chain-of-thought reasoning with next-token predictions to verify and generate solutions, although at a high computational cost. To balance these trade-offs, our DyVe framework merges the strengths of PRMs and GenRMs using Kahneman's dual system theory.
High-quality step-level supervision is crucial for training process verifiers, yet human annotations (e.g., PRM800k~\cite{Lightman2023LetsVS800k}) are prohibitively expensive. To avoid this, OmegaPRM~\cite{luo2024improve_omegaprm} employs a divide-and-conquer Monte Carlo Tree Search (MCTS) to generate annotations, although our experiments show that these labels are often noisy and weak. To address this issue, we adopt consensus filtering with an LLM-as-a-Judge~\cite{Gu2024ASOJudge} to eliminate unreliable samples~\cite{zhang2025lessons}, and further extend this approach with step-wise flagging, where a reasoning LLM conducts step-by-step analysis to identify steps that require System 2 verification.

\section{Method}

\subsection{Overview}
Dyve can assess the correctness of multi-step reasoning trace generated by a language model. Given a problem \(P\) and its reasoning steps \(\{s_1, s_2, \ldots, s_T\}\), Dyve sequentially verifies each step:
\[
r_t = \text{Dyve}(s_{1:t}; \theta)
\]
where the response \(r_t\), varying from 1 to 8192 tokens based on System 1 or System 2 usage, is parsed by \(\text{Parse}(\cdot)\) to yield a binary outcome. If \(\text{Parse}(r_t) = 0\), the process halts, returning the erroneous step index and intermediate generations; otherwise, verification proceeds to the next step.

\begin{figure}
    \centering
    \includegraphics[width=1.1\linewidth]{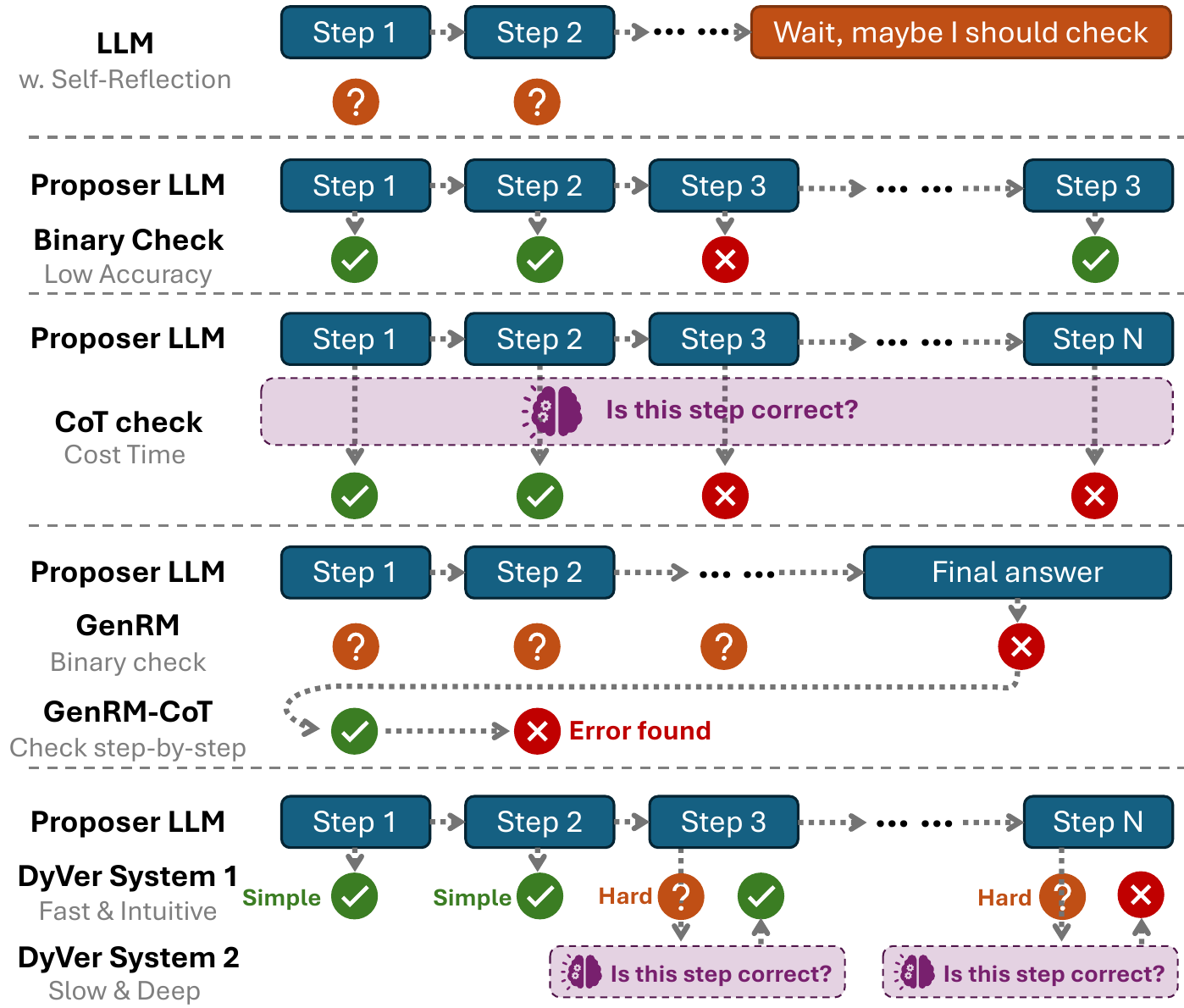}
    \caption{(1) LLM self-reflection is unreliable (2) Binary verification lacks depth, (3) Chain-of-Thought (CoT) verification is deeper but more expensive, (4) GenRM with CoT combines generation and verification without step-wise assessment, (5) Dyve, our proposed framework that dynamically combines fast System 1 and deep System 2 verification.}
    \label{fig:enter-label}
    \vspace{-15pt}
\end{figure}

\subsection{Step-wise Consensus-Filtered Process Supervision}
We introduce a novel step-wise consensus-filtered process supervision technique to enable adaptive verification within Dyve. The pipeline includes:

\paragraph{Queries Collection}
We gather query-response pairs from datasets like GSM8k~\cite{cobbe2021gsm8k} and MATH~\cite{Hendrycks2021MeasuringMP}, totaling 15K queries. 

\paragraph{Monte Carlo Rollouts Generation}
Using OmegaPRM~\cite{luo2024improve_omegaprm}, we generate 20 rollouts per query. We also gather open-souce PRM data from MathShepherd~\cite{wang2024mathshepherd} and RLHFlow, excluding PRM800k~\cite{Lightman2023LetsVS800k} to prevent data leakage, yielding approximately 1.2 million positive and negative rollouts with noisy labels.

\paragraph{Consensus Filtering with LLM-as-Judges}
We prompt DeepSeek V3 to verify the initial error steps identified by OmegaPRM. This filtering removes about 50\% of noisy rollouts. We then create a dataset of 117K high-quality examples by re-balancing the number of positive and negative step labels.

\paragraph{Step-Level Analysis with Reasoning LLMs}
A reasoning model performs step-by-step analysis on curated rollouts. Correct steps are marked with a “+” token, while uncertain steps undergo further detailed evaluation, ensuring alignment with high-quality reasoning traces.

\subsection{Training}
We train the {deepseek-ai/DeepSeek-R1-Distill-Qwen-14B} model using supervised fine-tuning on our curated dataset. This enables the model to learn rapid System 1 verification and comprehensive System 2 correction. The training objective minimizes the cross-entropy loss:

\begin{equation}
\mathcal{L}(\theta) = -\frac{1}{N} \sum_{i=1}^{N} \sum_{t=1}^{T^{(i)}} \log p_\theta \left( y_t^{(i)} \mid x^{(i)}, y_{<t}^{(i)} \right),
\end{equation}
where \(\theta\) indicates the model parameters, \(x^{(i)}\) is the input query, and \(y^{(i)}\) is the target label for the \(i\)-th example.

\begin{table*}[h]
\centering
\resizebox{\linewidth}{!}{
\begin{tabular}{llcccc}
\hline
Model                                               &                                        & \multicolumn{1}{l}{GSM8K} & \multicolumn{1}{l}{MATH} & \multicolumn{1}{l}{OlympiadBench} & \multicolumn{1}{l}{OmniMATH} \\ \hline
\multicolumn{1}{l|}{Qwen2.5-Math-7B-PRM}        & \multicolumn{1}{l|}{System1}           & 39.4$^*$                     & 52.2$^*$                     & 39.4$^*$                              & 33.1$^*$                         \\
\multicolumn{1}{l|}{Math-Shepherd-PRM-7B}           & \multicolumn{1}{l|}{System1}           & 47.9                      & 29.5                     & 24.8                              & 23.8                         \\
\multicolumn{1}{l|}{RLHFlow-PRM-Mistral-8B}         & \multicolumn{1}{l|}{System1}           & 50.4                      & 33.4                     & 13.8                              & 15.8                         \\
\multicolumn{1}{l|}{RLHFlow-PRM-Deepseek-8B}        & \multicolumn{1}{l|}{System1}           & 38.8                      & 33.8                     & 16.9                              & 16.9                         \\
\multicolumn{1}{l|}{Skywork-PRM-1.5B}               & \multicolumn{1}{l|}{System1}           & 59.0                      & 48.0                     & 19.3                              & 19.2                         \\
\multicolumn{1}{l|}{Skywork-PRM-7B}                 & \multicolumn{1}{l|}{System1}           & 64.1$^*$                       & 43.2$^*$                     & 16.2$^*$                              & 17.9$^*$                         \\ \hline
\multicolumn{1}{l|}{Llama-3.1-8B-Instruct}          & \multicolumn{1}{l|}{LLM-as-Judge}      & 27.5$^*$                      & 26.7$^*$                     & 18.5$^*$                              & 19.2$^*$                         \\
\multicolumn{1}{l|}{GPT-4o}                         & \multicolumn{1}{l|}{LLM-as-Judge}      & 61.9$^*$                      & 53.9$^*$                     & 48.3$^*$                              & 44.6$^*$                         \\
\multicolumn{1}{l|}{QwQ-32B-Preview}                        & \multicolumn{1}{l|}{LLM-as-Judge}      & 62.3$^*$                      & 52.7$^*$                     & 46.2$^*$                              & 43.9$^*$                         \\
\multicolumn{1}{l|}{DeepSeek-R1-Distill-Qwen-14B} & \multicolumn{1}{l|}{LLM-as-Judge}      & 67.3$^*$                      & 38.8$^*$                     & 29.9$^*$                              & 32.1$^*$                         \\ \hline
\multicolumn{1}{l|}{\textbf{Dyve 14B}}                  & \multicolumn{1}{l|}{System1 + System2} & \textbf{68.5}             & \textbf{58.3}            & \textbf{49.0}                     & \textbf{47.2}                \\ \hline
\end{tabular}
}
\caption{Performance comparison on ProcessBench. F1 scores, computed from accuracies on erroneous and correct samples, are reported for four benchmarks: GSM8K, MATH, OlympiadBench, and OmniMATH. Dyve 14B leverages a dual reasoning approach (fast System1 and slow System2) to achieve superior performance, with scores of 68.5, 58.3, 49.0, and 47.2, respectively, and it shows enhanced generalization on Olympiad-level mathematics. Models marked with a $^*$ are evaluated using our custom implementation to align with our experimental settings in the absence of an official evaluation script.}
\label{tab:value_head_prm}
\end{table*}

\section{Experiments}
To evaluate Dyve's capabilities, we conduct experiments in two main areas. First, we assess Dyve's ability to identify process errors. Second, we integrate Dyve with Proposer LLMs using a Best-of-N approach to evaluate its synergy within a reasoning framework. All experiments are conducted on 8 $\times$ NVIDIA A800-SXM4-80GB GPUs. Interested Readers may refer to Appendix ~\ref{sec:expSetup} for detailed experimental setup.

\subsection{Benchmarks}

\paragraph{ProcessBench}~\cite{zheng2024processbench} comprises four sets of test data derived from GSM8K~\cite{cobbe2021gsm8k}, MATH~\cite{Hendrycks2021MATH}, OlympiadBench~\cite{he2024olympiadbench}, and OmniMATH~\cite{Gao2024OmniMATHAU}. It includes 3,400 test cases, covering high-school to Olympiad-level math problems. Each case provides a step-by-step solution with error locations annotated by experts. \textit{Models are given \(s_{1:t}\), from the first to the last step, and must identify the earliest error or confirm that all steps are correct.} For each ProcessBench subset, we calculate the accuracies for erroneous and correct samples and compute their harmonic mean as the F1 score.

\paragraph{MATH-500} \cite{Lightman2023LetsVS800k} evaluates Dyve's integration with a Proposer LLM. We measure performance using maj@k and rm@k metrics as defined in \cite{yang2024qwen2.5math} and apply a Best-of-N decoding strategy. Due to inconsistent results from different evaluation tools, we manually verified all reported outcomes.

\subsection{Processbench}

\paragraph{Results and Analysis}
Dyve achieves the highest F1 scores across all benchmark subsets, outperforming all baselines. Despite being trained primarily on high-school and college-level mathematics, its dual reasoning system generalizes effectively to Olympiad-level problems. In contrast, LLM-as-Judge with DeepSeek-R1-Distill-Qwen-14B shows weaker performance on OlympiadBench and OmniMATH, indicating less reliable process error detection.

\begin{figure}[t]
    \centering
    \includegraphics[width=\linewidth]{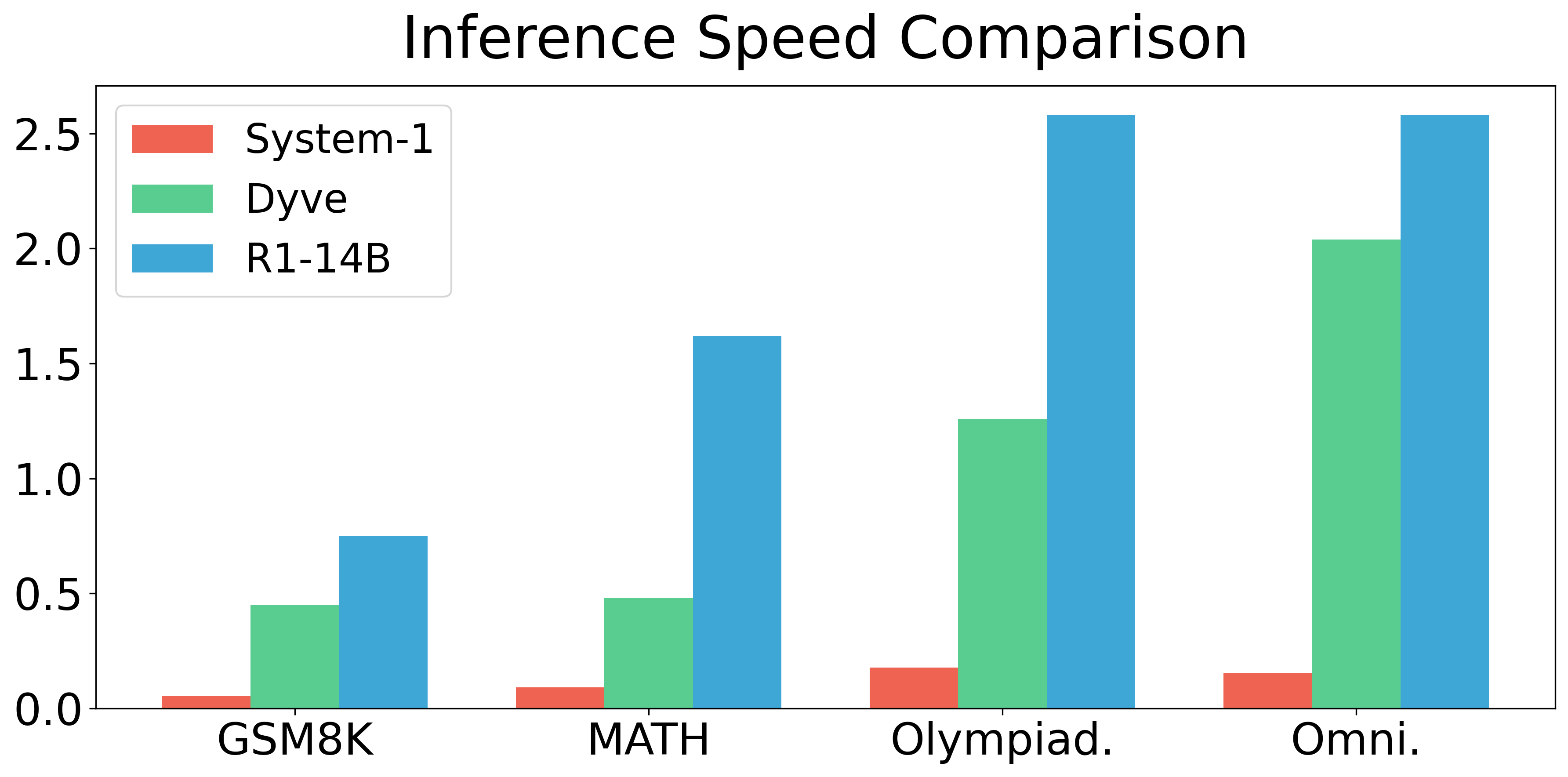}
    \caption{Inference speed comparison on ProcesBench, time per sample in seconds, for System-1, Dyve, and DeepSeek-R1-14B.}
    \label{fig:time}
    \vspace{10pt}
\end{figure}

\paragraph{Camparison on Inference Time}
According to Figure~\ref{fig:time}, the inference speed comparison in ProcesBench, highlights model efficiency. System-1 is the fastest, maintaining minimal latency. Dyve, slightly slower, balances speed and performance, excelling in complex datasets like OlympiadBench and OmniMATH. R1-14B has the longest inference times, suggesting a bottleneck for rapid processing. This analysis highlights Dyve's ability to deliver competitive performance with efficient inference times, making it well-suited for applications demanding both accuracy and speed.

\begin{figure}[t]
    \vspace{-10pt}
    \centering
    \includegraphics[width=\columnwidth]{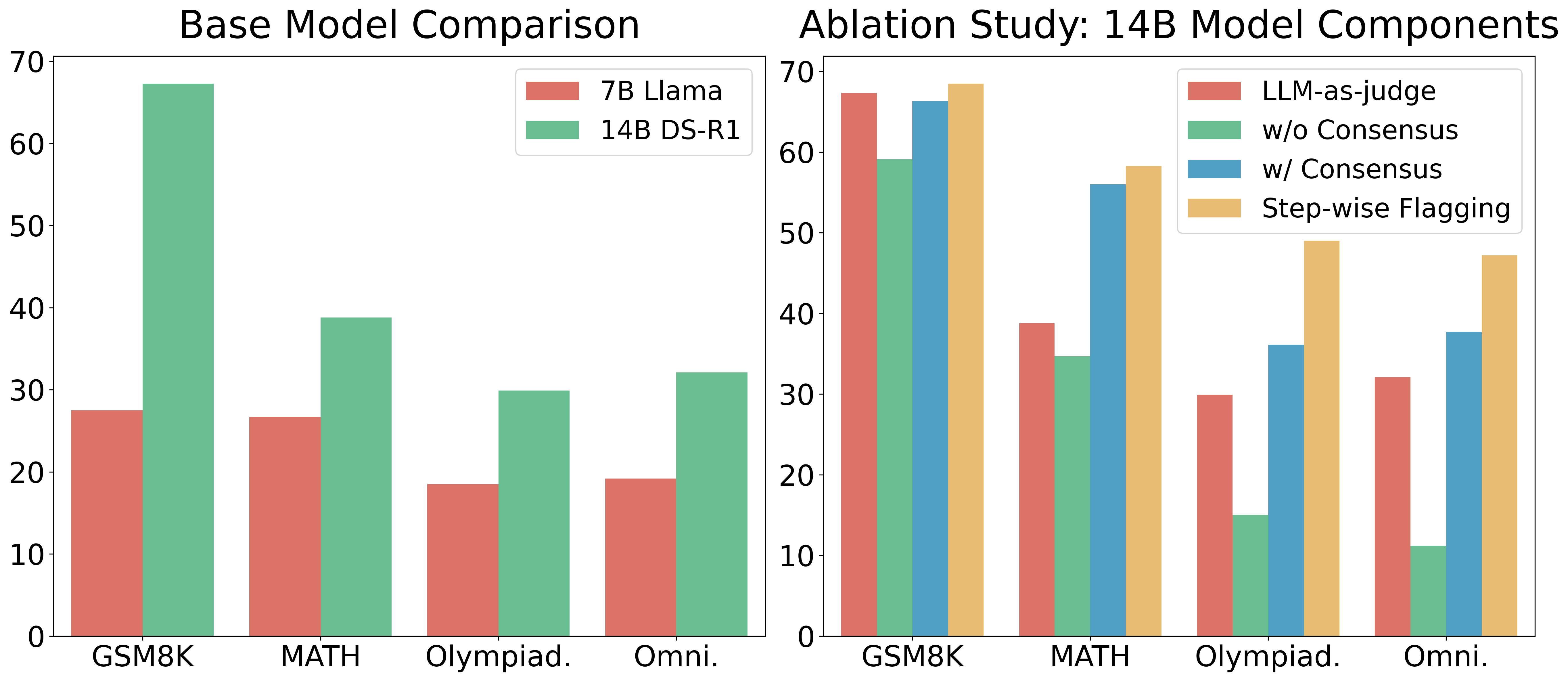}
    \caption{Impact of model choice and step-wise consensus filtering on performance across GSM8K, MATH, OlympiadBench, and OmniMATH. The figure illustrates improvements achieved through consensus filtering and step-wise flagging, highlighting the superior performance of the 14B reasoning model over the 7B Llama.}
    \label{fig:abl}
    \vspace{-15pt}
\end{figure}

\paragraph{Model Choice and Step-wise Consensus Filtering}
The ablation study in Figure~\ref{fig:abl} illustrates the impact of model selection and step-wise consensus filtering in ProcessBench. For Llama-3.1-8B-Instruct, consensus filtering significantly improves performance, boosting scores from 35.6 to 49.3 on GSM8K and from 28.3 to 40.2 on MATH. Similarly, DS-R1-Distill-Qwen-14B sees substantial gains, with MATH scores increasing from 34.7 to 56.0 and OmniMATH from 11.2 to 37.7. Step-wise flagging further amplifies performance, achieving scores of 68.5 on GSM8K and 58.3 on MATH. These results underscore the effectiveness of these techniques and highlight the superior reasoning capabilities of the 14B model compared to the 7B Llama, validating our choice of DeepSeek-R1-Distill-Qwen-14B.

\begin{figure}[h]
    \vspace{-20pt}
    \centering
    \includegraphics[width=1\columnwidth]{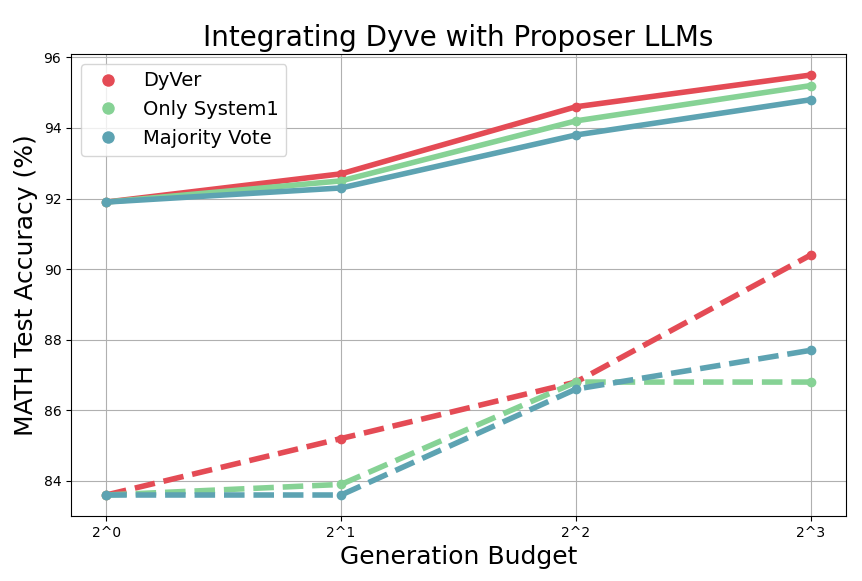}
    \caption{Comparison of Dyve, Dyve System1 and Majority Vote with different generation budget when integrating with Proposer LLMs (DeepSeek-R1-Distill-Qwen-14B as solid line, Qwen2.5-MATH-7B-Instruct as dotted line).}
    \label{fig:integarte_dyve}
    \vspace{-15pt}
\end{figure}

\subsection{Integrating Dyve with Proposer LLMs}
We integrate Dyve as a process verifier to assist Proposer LLMs (Qwen-Math-7B and Deepseek-R1-Distill-Qwen-14B) on MATH-500. For fairness, we compare three setups across Best-of-N (N = 1, 2, 4, 8) decoding settings: Dyve verification, System 1 only, and Majority Vote (no verification).

\vspace{-5pt}
\paragraph{Results and Analysis}
As shown in Figure \ref{fig:integarte_dyve}, Dyve's combination of fast and slow verification outperforms both Majority Voting and System 1 verification when integrated with Best-of-N decoding. When the generation budget is N = 8, Dyve with DeepSeek-R1-Distill-Qwen-14B achieves 95.5\% accuracy, while Dyve with Qwen2.5-MATH-7B-Instruct reaches 90.4\%, outperforming both baselines.
This demonstrates how our dual-system with fast and slow thinking, approach effectively guides Proposer LLMs to select more accurate reasoning paths, showcasing the synergy between the Dyve and proposer models.


\section{Conclusion}
Our study demonstrates Dyve's, with a dual reasoning approach, superior performance in mathematical reasoning verification. The consensus filtering and step-wise flagging significantly enhanced model accuracy and robustness. Ablation studies confirm the 14B model's advantages over smaller variants for complex reasoning tasks, establishing Dyve as an effective solution for precise and efficient error detection.

\label{submission}


\section{Broader Ethical Impact}

Our method is centered on rigorous verification of AI reasoning, ensuring each step is systematically validated for enhanced reliability and transparency. By exclusively using publicly available datasets under their proper licenses, we adhere to responsible research practices. We believe that improving verification in AI reasoning not only boosts system robustness but also exemplifies ethical AI development.

\section{Limitations}

While Dyve demonstrates strong performance, it shares several limitations common to verification-based systems. Its effectiveness naturally depends on the complexity of the reasoning tasks, and more intricate multi-step problems may require further adaptation or deeper analysis. In addition, although our consensus-filtered process supervision considerably enhances signal quality, a modest level of noise remains inherent in any automated estimation process. Finally, the overall performance is influenced by the quality and diversity of the training data, suggesting that further efforts in data curation and filtering could yield even more robust results. These aspects offer promising directions for future research.

\bibliography{custom}

\appendix
\section{Appendix}
\label{sec:appendix}

\subsection{Detailed Experiment Setup}
\label{sec:expSetup}

\paragraph{Training}
\subsection{Training Details}
Our model processes inputs with a maximum token length of 2048, ensuring robust contextual understanding. To further enhance efficiency, we employ Low-Rank Adaptation (LoRA) configured with a rank of 16, an alpha value of 16, and a dropout rate of 0.1. The training regimen spans three epochs, using a per-device batch size of 2 and leveraging gradient accumulation over 8 steps. The learning rate is set to \(2 \times 10^{-5}\) and a weight decay of 0.01 is applied. Training is executed with mixed precision (fp16), optimizing computational resources without sacrificing performance.

\paragraph{Inference}
During inference, our model leverages a multi-step reasoning process to evaluate each problem instance. The procedure begins by formulating a sequence of conversational prompts that encapsulate both the problem statement and its progressive steps. At each step, the Dyve model is queried via its custom chat interface, and the generated response is examined for specific response patterns — such as the presence of a "\(+\)" symbol signaling a correct evaluation. This iterative mechanism continues until a response fails to meet the designated correctness criteria, at which point the process halts. To ensure efficiency, the inference is executed concurrently using a pool of 32 parallel workers, processing various configurations from the ProcessBench dataset (including \texttt{gsm8k}, \texttt{math}, \texttt{olympiadbench}, and \texttt{omnimath}). For every evaluated problem, all intermediate responses (or generations) and the final step classification are recorded. These results are then systematically saved in JSON Lines format, facilitating subsequent analysis and serving as a robust foundation for further evaluation.

\subsection{Efficient Estimation of MCTS}
\label{sec:effmcts}

In this section, we detail our approach to efficiently utilize Monte Carlo Tree Search (MCTS) for sampling rollouts, which are crucial for training process-based verifiers.

\subsubsection*{Overview}

Our method leverages MCTS to construct a state-action tree representing detailed reasoning paths for a given question. This approach allows us to collect Process-based Reward Model (PRM) training examples by exploring various reasoning paths and identifying errors efficiently.

\subsubsection*{State-Action Tree Construction}

Each state \( s \) in the tree corresponds to a question and its preceding reasoning steps, with the root state being the question without any reasoning steps. An action \( a \) is a potential next step, and the state transition function is defined as \( s' = \text{Concatenate}(s, a) \). Each node \( s \) stores the visit count \( N(s) \), Monte Carlo estimation \( MC(s) \), and rollout value function \( Q(s, r) \).

\subsubsection*{MCTS Process}

\paragraph{Selection}

We maintain a pool of rollouts with \( 0 < MC(s) < 1 \). During selection, a rollout is chosen based on tree statistics using a variant of the PUCT algorithm:
\[
U(s) = c_{\text{puct}} \frac{\sqrt{\sum_i N(s_i)}}{1 + N(s)}
\]
This strategy initially favors rollouts with low visit counts, gradually shifting preference towards those with high rollout values.

\paragraph{Binary Search}

A binary search identifies the first error location in the selected rollout. Rollouts with \( 0 < MC(s) < 1 \) are added to the selection pool. The search process divides and examines rollouts to pinpoint errors, updating the tree with new states.

\paragraph{Maintenance}

After binary search, update the statistics \( N(s) \), \( MC(s) \), and \( Q(s, r) \). Increment \( N(s) \) for the selected state-action pair and update \( MC(s) \) and \( Q(s, r) \) based on the binary search results.

\paragraph{Tree Construction}

Repeat the above process to construct the state-action tree. The process ends when the search count reaches a predetermined limit or no additional rollouts are available.

\subsection{Finetuning Data Example}
\label{sec:DataExp}
\begin{figure*}[t]
\begin{minipage}{\textwidth}
\begin{tcolorbox}[
    colback=blue!5!white,
    colframe=blue!75!black,
    title=Example Data Sample,
    width=\textwidth,
    enhanced,
    boxrule=1pt,
    arc=4mm,
    auto outer arc,
    fonttitle=\bfseries
]
\footnotesize
\textbf{LLM Proposer:}
\begin{verbatim}
Step: Cowboy Mickey and cowgirl Minnie train horses. Mickey mounts six less than twice as 
many horses per day as Minnie, while Minnie mounts three more horses per day than there are days in 
a week. How many horses does Mickey mount per week?

Step 1: Minnie mounts 7+3=<<7+3=10>>10 horses per day.

Is this step correct? Answer with '+' for correct or '-' for incorrect.
\end{verbatim}

\textbf{DyVer:}
\begin{verbatim}
<think>
</think> +
\end{verbatim}

\textbf{LLM Proposer:}
\begin{verbatim}
Step 2: Mickey mounts 12-6=<<12-6=6>>6 horses per day.

Is this step correct? Answer with '+' for correct or '-' for incorrect.
\end{verbatim}

\textbf{DyVer:}
\begin{verbatim}
<think>
Let's verify Step 2 .... 

Mickey mounts six less than twice Minnie's count. Minnie mounts 10 horses, so Mickey should mount 
2*10 - 6 = 14. Step 2 is incorrect.
</think>
Answer: -
\end{verbatim}
\end{tcolorbox}
\end{minipage}
\end{figure*}

\end{document}